\documentclass{article}

\PassOptionsToPackage{numbers, compress}{natbib}
\usepackage[preprint]{neurips_2020}

\usepackage[utf8]{inputenc} % allow utf-8 input
\usepackage[T1]{fontenc}    % use 8-bit T1 fonts
\usepackage{hyperref}       % hyperlinks
\usepackage{url}            % simple URL typesetting
\usepackage{booktabs}       % professional-quality tables
\usepackage{amsfonts}       % blackboard math symbols
\usepackage{nicefrac}       % compact symbols for 1/2, etc.
\usepackage{microtype}      % microtypography
\usepackage{graphicx}
\usepackage{caption}
\usepackage{amsmath}

\usepackage{xcolor}

\title{Graph Representation Learning Network via Adaptive Sampling}

\author{%
  Anderson de Andrade \\
  Wattpad \\
  Toronto, Ontario, Canada \\
  \texttt{anderson@wattpad.com} \\
  \And
  Chen Liu \\
  Wattpad \\
  Toronto, Ontario, Canada \\
  \texttt{cecilia@wattpad.com} \\
}

\begin{document}

\maketitle

\begin{abstract}
Graph Attention Network (GAT) and GraphSAGE are neural network architectures that operate on graph-structured data and have been widely studied for link prediction and node classification. One challenge raised by GraphSAGE is how to smartly combine neighbour features based on graph structure. GAT handles this problem through attention, however the challenge with GAT is its scalability over large and dense graphs. In this work, we proposed a new architecture to address these issues that is more efficient and is capable of incorporating different edge type information. It generates node representations by attending to neighbours sampled from weighted multi-step transition probabilities. We conduct experiments on both transductive and inductive settings. Experiments achieved comparable or better results on several graph benchmarks, including the Cora, Citeseer, Pubmed, PPI, Twitter, and YouTube datasets.
\end{abstract}

\section{Introduction}
Graphs are a versatile and succinct way to describe entities through their relationships. The information contained in many knowledge graphs (KG) has been used in several machine learning tasks in natural language understanding \cite{Peters2019KnowledgeEC}, computer vision \cite{li2017situation}, and recommendation systems \cite{wang2019explainable}. The same information can also be used to expand the graph itself via node classification \cite{grover2016node2vec}, clustering \cite{perozzi2014deepwalk}, or link prediction tasks \cite{kazemi2018simple}, in both transductive and inductive settings \cite{kipf2016semi}.

Recent graph models have focused on learning dense representations that capture the properties of a node and its neighbours. One class of methods generates spectral representation for nodes \cite{bruna2013spectral, defferrard2016convolutional}. The rigidity of this approach may reduce the adaptability of a model to graphs with structural differences.

Model architectures that reduce the neighbourhood of a node have used pooling \cite{hamilton2017inductive}, convolutions \cite{duvenaud2015convolutional}, recurrent neural networks (RNN) \cite{li2015gated}, and attention \cite{velivckovic2017graph}. These approaches often require many computationally-expensive message-passing iterations to generate representations for the neighbours of a target node. Sparse Graph Attention Networks (SGAT) \cite{ye2019sparse} was proposed to address this inefficiency by producing an edge-sparsified graph. However, it may neglect the importance of local structure when graphs are large or have multiple edge types. Other methods have used objective functions to predict whether a node belongs to a neighbourhood \cite{perozzi2014deepwalk, kazemi2018simple}, for example by using noise-contrastive estimation \cite{gutmann2012noise}. However, incorporating additional training objectives into a downstream task can be difficult to optimize, leading to a multi-step training process.

In this work, we present a graph network architecture (GATAS) that can be easily integrated into any model and supports general graph types, such as: cyclic, directed, and heterogeneous graphs. The method uses a self-attention mechanism over a multi-step neighbourhood sample, where the transition probability of a neighbour at a given step is parameterized.

%Attention mechanisms allow for variable size sequences that can be combined selectively. It has produced state-of-the-art performance over RNNs and convolutions \cite{vaswani2017attention}. The attention mechanism offers the dynamism required for complex downstream tasks. Using neighbour samples allows us to configure the amount of overhead this architecture adds. Attending to neighbours that are multiple steps away from a target node, incorporating the directions and edge types between them, results in expressive node representations that do not require multiple iterations to generate intermediate neighbour representations, limiting the scope of the nodes required by a single node.

We evaluate the proposed method in node classification tasks using the Cora, Citeseer and Pubmed citation networks in a transductive setting, and on a protein to protein interaction (PPI) dataset in an inductive setting. We also evaluate the method on a link prediction task using a Twitter and YouTube dataset. Results show that the proposed graph network can achieve better or comparable performance to state-of-the-art architectures.

\section{Related Work}

The proposed architecture is related to GraphSAGE \cite{hamilton2017inductive}, which also reduces neighbour representations from fixed-size samples. Instead of aggregating uniformly-sampled 1-hop neighbours at each depth, we propose a single reduction of multi-step neighbours sampled from parameterized transition probabilities. Such parameterization is akin to the Graph Attention Model \cite{abu2018watch}, where trainable depth coefficients scale the transition probabilities from each step. Thus, the model can choose the depth of the neighbourhood samples. We further extend this approach to transition probabilities that account for paths with heterogeneous edge types.

We use an attention mechanism similar to the one in Graph Attention Networks (GAT) \cite{velivckovic2017graph}. While GAT reduces immediate neighbours iteratively to explore the graph structure in a breadth-first approach that processes all nodes and edges at each step, our method uses multi-step neighbourhood samples to explore the graph structure. Our method also allows each neighbour to have a different representation for the target node, rather than using a single representation as in GAT.

MoNet \cite{monti2017geometric} generalizes many graph convolutional networks (GCN) as attention mechanisms. More recently, the edge-enhanced graph neural network framework (EGNN) \cite{gong2019exploiting} consolidates GCNs and GAT. In MoNet, the attention coefficient function only uses the structure of the nodes. In addition, our model employs node representations to generate attention weights.

Other approaches use recurrent neural networks \cite{scarselli2008graph, li2015gated} to reduce path information between neighbours and generate node representations. The propagation algorithm in  Gated Graph Neural Networks  (GG-NNs) \cite{li2015gated} reduces neighbours one step at a time, in a breadth-first fashion. In contrast, we use a depth-first approach where a reduction operation is applied across edges of a path.

\section{Model}

%We start describing our model in Section \ref{preliminaries} by defining the graph notation, and some necessary edge type and edge modifications. We define edge type paths, which will help us define our neighbour representations and transition probabilities. In Section \ref{neighbours} we define how to transform neighbour representations given a path between the neighbour and the target node. In Sections \ref{transitions} and \ref{neighbours} a probability tensor is defined so edge type paths between nodes can be sampled using learnable probability distributions. Finally, Section \ref{attention} defines the attention mechanism that creates node representations that capture structure information. Figure \ref{illustrations} illustrates different aspects of the proposed method. We analyze the complexity of the model in Section \ref{complexity} and compare against related work.

\subsection{Preliminaries} \label{preliminaries}

An initial graph is defined as $\mathcal{G} = (\mathcal{V}, \mathcal{R}, \mathcal{Z})$, where $\mathcal{V} = \{v_1,\allowbreak ...,\allowbreak v_{|\mathcal{V}|}\}$ is a set of nodes or vertices, $\mathcal{R} = \{r_1, ..., r_{|\mathcal{R}|}\}$ is a set of edge types or relations, and $\mathcal{Z} = \{(v_i, r_k, v_j) \mid v_i, v_j \in \mathcal{V}; r_k \in \mathcal{R}\}$ is a set of triplets. Directed graphs are represented by having one edge type for each direction so that $(v_i, r^+, v_j)\allowbreak \in \mathcal{Z} \iff (v_j, r^-, v_i) \in \mathcal{Z}$. To be able to incorporate information about the target node, the graph $\mathcal{G}$ is augmented with self-loops using a new edge type $r_{\emptyset}$. Thus, a new set of edge types $\tilde{\mathcal{R}} = \{r_{\emptyset}\} \cup \mathcal{R}$ is created such that $\tilde{r}_1 = r_{\emptyset}$, and a new set of triplets $\tilde{\mathcal{Z}} = \{(v_i, r_{\emptyset}, v_i) \mid v_i \in \mathcal{V}\} \cup \{(v_i, r_{k + 1}, v_j) \mid (v_i, r_k, v_j) \in \mathcal{Z}\}$ conforms a new graph $\tilde{\mathcal{G}} = (\mathcal{V}, \tilde{\mathcal{R}}, \tilde{\mathcal{Z}})$.

An edge type path between nodes $v_i$ and $v_j$ is defined as $E^{(i, j)} = (a_t \mid \tilde{r}_{a_t} \in \tilde{\mathcal{R}})_{t=1}^{\leq C}$, where $C \geq 1$ is the maximum number of steps considered. The number of all possible edge type paths is given by $M = \sum_{t=1}^C |\tilde{\mathcal{R}}|^t$. The set of all possible edge type paths is defined as $\mathcal{E} = \{E_1,\allowbreak ...,\allowbreak E_M\}$. The edge type sequences in the set are level-ordered so that $\forall m \in \{1, ..., M - 1\}, |E_m| \le |E_{m + 1}|$, and so that $\forall m, n \in \{(m, n) : |E_m| = |E_n|, m < n\}, \forall i \in \{1,...,|E_{m, i}|\}, E_{m, i} \le E_{n, i}$. As an example, if $\tilde{\mathcal{R}} = \{A, B\}$ and $C = 2$, then $\mathcal{E}$ corresponds to: $\{(A),\allowbreak (B),\allowbreak (A, A),\allowbreak (A, B),\allowbreak (B, A),\allowbreak (B, B)\}$. The subset of relation paths connecting nodes $v_i \to v_j$ is defined as $\mathcal{E}^{(i, j)} = \{E^{(i,j)}_m | E_m \in \mathcal{E}$\}, where $\mathcal{E}^{(i, j)} = \{E_1\} \Rightarrow i = j$ represents the extraneous self-loops.

\subsection{Neighbour Representations} \label{neighbours}

\begin{figure*}
  \includegraphics[width=\linewidth]{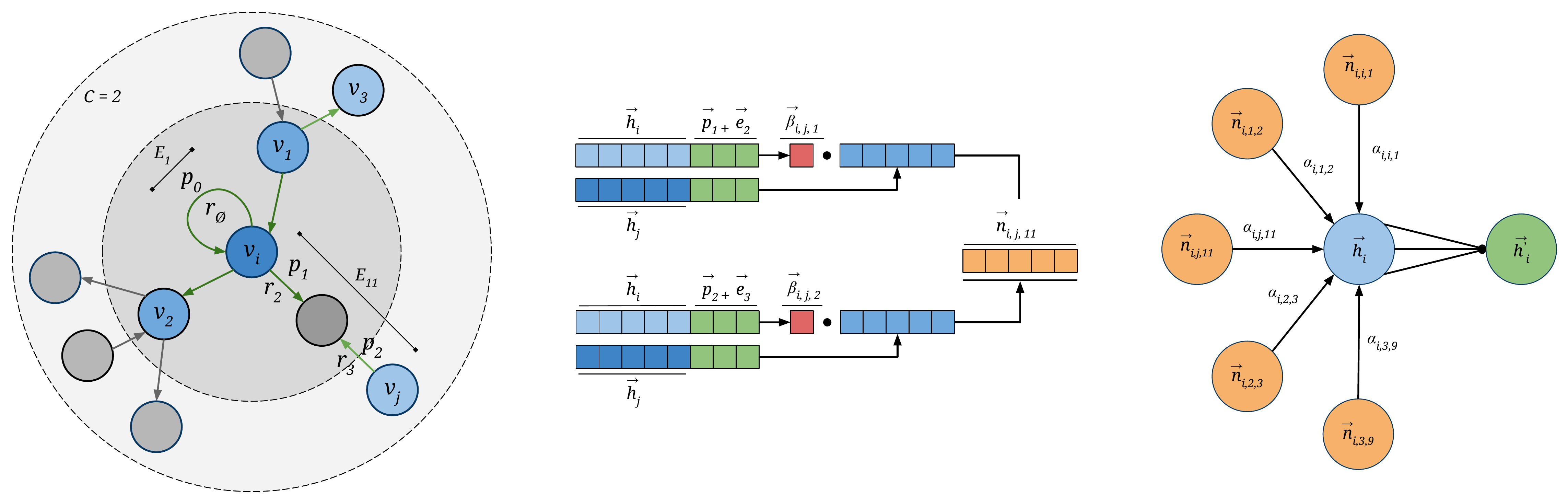}
 % \captionsetup{textfont=normalfont}
  \caption{\textbf{Left:} An illustration of the multi-step sampling technique with relevant notation presented in this work. Blue nodes represent sampled nodes. The edge type path $E_{11}$ is detailed. \textbf{Center:} The attention mechanism for $E_{11} \in \mathcal{E}^{(i, j)}$ that generates the neighbour representation $\vec{n}_{i,j,11}$. \textbf{Right:} The attention mechanism by target node $v_i$. Neighbour representations are aggregated according to the attention weights.}
  %\Description{Model Illustrations}
  \label{illustrations}
\end{figure*}

Graph relations in $\tilde{\mathcal{R}}$ are represented by trainable vectors $\mathbf{e} = \{\vec{e}_1,...,\vec{e}_{|\tilde{\mathcal{R}}|}\}, \vec{e} \in \mathbb{R}^D$. To reflect the position in a path, the edge representations can be infused with information that reflects its position in a path. Following \cite{vaswani2017attention}, we assign a sinusoid of a different wavelength to each dimension, each position being a different point:
\begin{equation*}
  \vec{p}_{t, 2i} = \sin (t / 10000^{2i/ D}),
  \textrm{ }
  \vec{p}_{t, 2i + 1} = \cos (t / 10000^{2i/ D})
\end{equation*}
where $t: 0 \leq t \leq C$ is the position in an edge type path and $i: 0 \leq i < D / 2$ is the index of the dimension.

We represent nodes as a set of vectors $\vec{\mathbf{h}} = \{\vec{h}_1, ..., \vec{h}_{|V|}\}, \vec{h}_i \in \mathbb{R}^{F + R}$. Each vector is defined by $\vec{h}_i = [\vec{l}_i \| \vec{b}_i]$, where $\vec{l}_i \in \mathbb{R}^{R}$ are trainable embedding representations and $\vec{b}_i \in \mathbb{R}^{F}; \vec{b}_i \in \vec{\mathbf{b}} = \{ \vec{b}_1, ..., \vec{b}_{|\mathcal{V}|} \}$ represents the features of the node. The learnable embedding representations can capture information to support a neighbourhood sample.

Given an edge type path $E_m \in \mathcal{E}^{(i, j)} | i \neq j$, we generate a neighbour representation $\vec{n}_{i, j, m}$ using attention over transformed neighbour representations for each edge type in $E_m$:
\begin{equation}
  \vec{n}_{i, j, m} =
  \sum_{t = 1}^{|E_m|}
  \beta_{i, m, t}\;
  z([\vec{h}_j \| \vec{e}_{E_{m, t}} \! + \vec{p}_t]),
  \textrm{ }
  \beta_{i, m, t} = \frac
  {\exp(f([\vec{h}_i \| \vec{e}_{E_{m, t}} \! + \vec{p}_t]))}
  {\sum_{x = 1}^{|E_m|} \exp(f([\vec{h}_i \| \vec{e}_{E_{m, x}} \! + \vec{p}_x]]))}
\end{equation}
where $z: \mathbb{R}^{F + R + D} \to \mathbb{R}^{F'}$ and $f: \mathbb{R}^{F + R + D} \to \mathbb{R}$ are two different learnable transformations. The transformation given by $z$ allows neighbour representations to be different according to the edge type in the path. For the self-loop edge type path $E_1$, we set $\vec{n}_{i, j, 1} = z([\vec{h}_i \| \vec{e}_1 \! + \vec{p}_0])$.

\subsection{Transition Tensors} \label{transitions}

%We define transition probability distributions for neighbours and their possible edge type paths within $[1, C]$ steps. These are used in Section \ref{sampling} to sample neighbourhoods. Corresponding edge type paths of these neighbours are also used to generate their representations in Section \ref{neighbours}.

We define transition probability distributions for neighbours and their possible edge type paths within $[1, C]$ steps. When there are multiple edges types connecting two nodes, their transition probabilities split. Thus, when computing transition probabilities for random walks starting at $v_i$, it is necessary to track the probability of an edge type path $E_m \in \mathcal{E}^{(i, j)}$ for each destination vertex $v_j$, effectively computing $P(v_j, \mathcal{E} | v_i)$. Also, when performing random walks from a starting node $v_i$, we break cycles by disallowing transitions to nodes already visited in previous steps. This reduces $\mathcal{E}^{(i, j)}$ to the set of shortest $v_i \to v_j$ edge type paths possible. 

Let $\mathbf{A} \in \{0, 1\}^{|\mathcal{V}| \times |\mathcal{V}| \times M}$ be a sparse adjacency tensor for $\tilde{\mathcal{G}}$, where:
\begin{equation*}
  A_{i, j, m} =
  \begin{cases}
    1, & \text{if }
    (v_i, \tilde{r}_m, v_j) \in \tilde{\mathcal{Z}} \wedge 1 < m \leq |\tilde{\mathcal{R}}| \\ 
    0, & \text{otherwise}
  \end{cases}
\end{equation*}
An initial transition tensor $\mathbf{T}^{(1)} \in \mathbb{R}^{|\mathcal{V}| \times |\mathcal{V}| \times M}$ can be computed by normalizing the $\mathbf{A}_{v, *, *}$ matrices to sum to one, when applying the function $\Psi : \mathbb{R}^{|\mathcal{V}| \times |\mathcal{V}| \times M} \to \mathbb{R}^{|\mathcal{V}| \times |\mathcal{V}| \times M}$:
\begin{equation*}
  \label{normalizer}
	\Psi(\mathbf{Z})_{i, j, m} = \frac
	{Z_{i, j, m}}
	{\sum_{x=1}^{|\mathcal{V}|} \sum_{y=1}^{M} Z_{i, x, y}}
\end{equation*}
Using the order in $\mathcal{E}$ to obtain the probabilities for specific edge type paths, the unnormalized sparse transition tensor $\tilde{\mathbf{T}}^{(t)}$ for steps $1 < t \leq C$ can be computed as follows:
\begin{equation}
  \label{t_unnormalized}
  \tilde{T}_{i, j, m}^{(t)} =
  \begin{cases}
    \displaystyle \sum_{x = 1}^{|\mathcal{V}|}
    T_{i, x, \phi(m)}^{(1)}
    T_{i, x, \delta(m)}^{(t - 1)},
    & \text{if }
    \displaystyle \sum_{y = 1}^{M}
    T^{(l)}_{i, j, y}\!= 0,
    \forall l \in \{1,...,t - 1\} \\

    0, & \text{otherwise}
  \end{cases}
\end{equation}
where $\phi(m) = (m - 1) \bmod |\tilde{\mathcal{R}}| + 1$ specifies the edge type $\tilde{r}_k$ for the last step in path $E_m$, and $\delta(m) = \lfloor (m - 1) / |\tilde{\mathcal{R}}| \rfloor$ is the edge type path $E_n$ without the last step in $E_m$. As an example, if $E_m$ is the sequence of relation indices that corresponds to $\{A, B, C\}$, then $E_{\delta(m)}$ corresponds to $\{A, B\}$, and $E_{\phi(m)}$ corresponds to $C$.

The conditional function in Equation \ref{t_unnormalized} sets the transition probability to zero if the node $v_j$ can be reached from $v_i$ in a previous step $t$. This procedure effectively breaks cycles and allows only the most relevant and shortest edge type paths to be sampled. A normalized transition tensor is obtained by $\mathbf{T}^{(t)} = \Psi(\mathbf{\tilde{T}}^{(t)})$. 

\subsection{Neighbourhood Sampling} \label{sampling}

When considering a neighbourhood $N_{i} = \{(j, m) \mid v_j \in \mathcal{V}, E_m \in \mathcal{E}^{(i, j)}\}$, it can be relevant to attend to nodes beyond the first degree neighbourhood. However, as the number of hops between nodes increases, their relationship weakens. It can also be prohibitive to attend to all nodes within $C$ hops, as the neighbourhood size $|N_i|$ grows proportionally with $C \times \mathbb{E}_{v \in |\mathcal{V}|}[\mathrm{degree}(v)]^C$.

To overcome these complications, we create a fixed-size neighbourhood sample from an adjustable transition tensor $\mathbf{P} \in \mathbb{R}^{|\mathcal{V}| \times |\mathcal{V}| \times M}$. Similar to the work in \cite{abu2018watch}, we obtain neighbour probabilities by a linear combination of random walk transition tensors for each step $k$, with learnable coefficients $\mathbf{q}$:
\begin{equation*}
	\mathbf{P} = \sum_{t=0}^{C} q_t \mathbf{T}^{(t)}
\end{equation*}
where  $(q_0, ..., q_C) = \mathrm{softmax}(\tilde{q}_0, ..., \tilde{q}_C)$ and $\tilde{\mathbf{q}} \in \mathbb{R}^{C + 1}$ is a vector of unbounded parameters. $\mathbf{T}^{(0)}$ corresponds to a transition tensor for the added self-loops:
\begin{equation*}
  T_{i, j, m}^{(0)} =
  \begin{cases}
    1, & \text{if }
    (v_i, r_{\emptyset}, v_j) \in \tilde{\mathcal{Z}}
    \wedge m = 1 \\ 
    0, & \text{otherwise}
  \end{cases}
\end{equation*}

Depending on the task and graph, the model can control the scope of the neighbourhood by adjusting these coefficients through backpropagation.

To generate a neighbourhood $N_i$ for $v_i \in \mathcal{V}$ we sample without replacement from $\mathbf{P}$ so that $N_i = \{(j, m) \mid (j, m) \sim \mathbf{P}_{i, *, *} \}^S$, where $S$ is the maximum size for a neighbourhood sample.

\subsection{Node Representations} \label{nodes}

Given a neighbourhood $N_i$ for node $v_i \in V$ and a transition tensor $\mathbf{P}$, we apply an attention mechanism with attention coefficients given by:
\begin{equation}
	\alpha^{(k)}_{i, j, m} = \frac
	{\exp[g^{(k)}([\vec{h}_i \| \vec{n}_{i,j,m}]) + \ln(P_{i, j, m})]}
	{\sum_{(x, y) \in N_i} \exp[g^{(k)}([\vec{h}_i \| \vec{n}_{i, x, y}]) + \ln(P_{i, x, y})]}
\end{equation}
where $g^{(k)}: \mathbb{R}^{F'} \to \mathbb{R}$ is a learnable transformation. The logits produced by $g^{(k)}$ are scaled by the transition probabilities, exerting the importance of the neighbour, and allowing the coefficients $\mathbf{q}$ to be trained. We concatenate multi-head attention layers to create a new node representation $\vec{h}'_i$:
\begin{equation}
  \vec{h}'_i =
  \Bigg\|_{k=1}^K
  \sigma
  \left(
    \sum_{(j, m) \in N_i}
    \alpha_{i, j, m}^{(k)} \;
    d^{(k)}(\vec{n}_{i, j, m})
  \right)
\end{equation}
where $d^{(k)}: \mathbb{R}^{F'} \to \mathbb{R}^{F''}$ is another learnable transformation, and $\sigma: \mathbb{R} \to \mathbb{R}$ is a non-linear activation function, such as the ELU \cite{clevert2015fast} function. The transformation allows relevant information for the node to be selected.

\subsection{Algorithmic Complexity} \label{complexity}

The complexity of generating node representations with the proposed algorithm (GATAS) is governed by $\mathcal{O}(C \times S \times B \times F_{max})$, where $B$ is the batch size and $F_{max} = \max(F + R + D, F', F'')$. GraphSAGE \cite{hamilton2017inductive} has a similar complexity of $\mathcal{O}(S^C \times B \times F)$. GAT \cite{velivckovic2017graph} on the other hand, has a complexity that is independent of the batch size but processes all nodes and edges. It is given by $\mathcal{O}(C \times (|\mathcal{V}| \times F + |\tilde{\mathcal{Z}}|))$, where $C$ is the number of layers that controls depth. For downstream tasks where only a small subset of nodes are actually used, the overhead complexity of GAT can be overwhelming. Generalizing, our model is more efficient when $|\mathcal{V}| + |\tilde{\mathcal{Z}}| > B \times S$.

\section{Evaluation}

We evaluate the performance of GATAS using node classification tasks in transductive and inductive settings. To evaluate the performance of the proposed attention mechanism over heterogeneous multi-step neighbours, we rely on a multi-class link prediction task.
% In a transductive setting the model might access nodes that are classified during testing. In an inductive setting the model trains on different graphs and has never seen any of the test graphs during training.

For the transductive learning experiments we compare against GAT \cite{velivckovic2017graph} and some of the approaches specified in \cite{kipf2016semi}, including a CNN approach that uses Chebyshev approximations of the graph eigendecomposition \cite{defferrard2016convolutional}, the Graph Convolutional Network (GCN) \cite{kipf2016semi}, MoNet \cite{monti2017geometric}, and the Sparse Graph Attention Network (SGAT) \cite{ye2019sparse}. We also benchmark against a multi-layer perceptron (MLP) that classifies nodes only using its features without any graph structure.

For the inductive experiments we compare once again against GAT \cite{velivckovic2017graph} and SGAT \cite{ye2019sparse}. We also compare against GraphSAGE \cite{hamilton2017inductive}, a method that aggregates node representations from fixed-size neighbourhood samples, using different methods such as LSTMs and max-pooling.

GATAS is capable of utilizing edge information, which we consider to be an important advantage. Hence we also conducted link prediction experiments on multiplex heterogeneous network datasets against some of the state-of-the art models, namely GATNE\cite{2019GATNE}, MNE\cite{2018mne}, and MVE\cite{qu2017mve}. GATNE creates multiple representations for a node under different edge type graphs, aggregates these individual views using reduction operations similar to GraphSAGE, and combines these node representations using attention.

\subsection{Datasets}

For the transductive node classification tasks we use three standard citation network datasets: Cora, Citeseer, and Pubmed \cite{sen2008collective}. In these datasets, each node corresponds to a publication and undirected edges represent citations. Training sets contain 20 nodes per class. The validation and test sets have 500 and 1000 unseen nodes respectively.
%The Cora dataset has 2708 nodes, 5429 edges, 7 classes, 1433 features per node, and on average 3.9 neighbours per node. The Citeseer dataset has 3327 nodes, 4732 edges, 6 classes, 3703 features per node and on average 2.7 neighbours per node. The Pubmed dataset has 19717 nodes, 44338 edges, 3 classes, 500 features per node, and on average 4.4 neighbours per node.

For the inductive node classification experiments, we use the protein interaction dataset (PPI) in \cite{hamilton2017inductive}. The dataset has multiple graphs, where each node is a protein, and undirected edges represent an interaction between them. Each graph corresponds to a different type of interaction between proteins. 20 graphs are used for training, 2 for validation and another 2 for testing.
%Each protein has 50 features that are composed of positional gene sets, motif gene sets and immunological signatures. A protein can have 121 possible labels simultaneously, and they correspond to a node set from the gene ontology collected from the Molecular Signatures Database \cite{subramanian2005gene}. The protein interaction dataset has 56944 nodes, 818716 edges, and on average 28.3 neighbours per node.

For the link prediction task, we use the heterogeneous Higgs Twitter Dataset\footnote{\url{http://snap.stanford.edu/data/higgs-twitter.html}} \cite{2013twitter}. It is made up of four directional relationships between more than 450,000 Twitter users. We also use a multiplex bidirectional network dataset that consists of five types of interactions between 15,088 YouTube users \cite{tang2009uncovering, tang2009uncoverning2}. Using the dataset splits provided by the authors of GATNE\footnote{\url{http://github.com/thudm/gatne}}, we work with subsets of 10,000 and 2,000 nodes for Twitter and YouTube respectively, reserving 5\% and 10\% of the edges for validation and testing. Each split is augmented with the same amount of non-existing edges, that are used as negative samples.

Detailed statistics for these datasets are summarized in Table \ref{datastats} in Supplementary Materials.

\subsection{Experiment Setup}
\label{experiment_setup}

Node features are normalized using layer normalization \cite{ba2016layer}. These features are then passed through a single dense layer to obtain the input features $\vec{\mathbf{b}}$. The Twitter dataset does not provide node features so $\vec{\mathbf{h}} = \vec{\mathbf{l}}$. The inductive node classification task does not use learnable node embeddings so $\vec{\mathbf{h}} = \vec{\mathbf{b}}$. We define $f(\cdot)$ as a linear transformation, $g^{(k)}(\cdot)$ as a two-layer neural network with a non-linear hidden layer and a linear output layer, and $z(\cdot)$ and $d^{(k)}(\cdot)$ as one-layer non-linear neural networks. Non-linear layers use ELU \cite{clevert2015fast} activation functions.

For all models we set $F, F', F'' = 50$. We experimented with learnable node embeddings and edge type embedding sizes of 10 and 50 for the transductive node classification and link prediction tasks respectively. We use an edge type embedding size of 5 for the inductive tasks. The transductive tasks have 8 attention heads, while the other tasks have 10 heads.

For the transductive node classification tasks, the output layer is directly connected to the concatenated attention heads. For the inductive task, the concatenated attention heads are passed through 2 non-linear layers before going through the output layer. In the link prediction task, the concatenated attention heads are passed through a non-linear layer and a pair of corresponding node representations are concatenated before they pass through 2 non-linear layers and an output layer. All these hidden layers have a size of 256.

The optimization objective is the multi-class or multi-label cross-entropy, depending on the task. It is minimized by the Nadam SGD optimizer \cite{dozat2016incorporating}. The validation set is used for early stopping and hyper-parameter tuning. Since the training sets for the transductive node classification tasks are very small, it is crucial to add noise to the model inputs to prevent overfitting. We mask out input features with 0.9 probability, and apply Dropout \cite{srivastava2014dropout} with 0.5 probability to the attention coefficients and resulting representations. We also add $L_2$ regularization with $\lambda = 0.05$.

In the node classification and link prediction tasks, neighbourhood candidates can be at most 3 and 2 steps $C$ away from the target node respectively. The unnormalized transition coefficients are initialized with a non-linear decay given by $\tilde{q}_t = -t /\ln(C + 1)$. To accommodate for an inductive setting, edges across graphs in the protein interaction dataset are treated as the same type and use the same edge type representations. In the link prediction task we reuse the node representations during test time and rely on the neighbours given by the edges in the training set.

The experiment parameters are summarized in Table \ref{parameters} in Supplementary Materials. In the transductive node classification experiments, the architecture hyper-parameters were optimized on the Cora dataset and are reused for Citeseer and Pubmed. A single experiment can be run on a V100 GPU under 12 hours. Implementation code is available on GitHub \footnote{\url{http://github.com/wattpad/gatas}}.

\begin{table*}
  \caption{Node Classification Results}
  \label{results}
  \begin{tabular*}{\linewidth}{l@{\extracolsep{\fill}}rrrr}
    \toprule
    & \multicolumn{3}{c}{Transductive (Accuracy \%)} & \multicolumn{1}{c}{Inductive (Micro-F1)} \\
    \midrule
    Model & Cora & Citeseer & Pubmed & PPI \\
    \midrule
    MLP & $55.1\%$ & $46.5\%$ & $71.4\%$ & $0.422$ \\
    Chebyshev \cite{defferrard2016convolutional} & $81.2\%$ & $69.8\%$ & $74.4\%$ & --- \\
    GCN \cite{kipf2016semi} & $81.5\%$ & $70.3\%$ & $79.0\%$ & --- \\
    MoNet \cite{monti2017geometric} & $81.7\% \pm 0.5\%$ & --- & $78.8\% \pm 0.3\%$ & --- \\
    \midrule
    GraphSAGE \cite{hamilton2017inductive} & --- & --- & --- & 0.768 \\
    GAT \cite{velivckovic2017graph} & $83.0\% \pm 0.7\%$ & $ \textbf{72.5\%} \pm 0.7\%$ & $ \textbf{79.0\%} \pm 0.3\%$ & $0.973 \pm 0.002$ \\
    SGAT$^*$ \cite{ye2019sparse} & \textbf{84.2\%} & 68.2\% & 77.6\% & 0.966\\
    % EGNN \cite{gong2019exploiting} & $83.4\% \pm 0.3\% ^+$  &  $70.6\% \pm  0.3\% ^+$ &  --- &  ---  \\
    \midrule
    GATAS$_{10}$ & $80.2\% \pm 1.1\%$ & $69.4\% \pm 1.3\%$ & $76.1\% \pm 0.8\%$ & $0.818 \pm 0.015$ \\
    GATAS$_{100}$ & $82.3\% \pm 0.9\%$ & $69.6\% \pm 1.1\%$ & $78.4\% \pm 0.6\%$ & $0.981 \pm 0.002$ \\
    GATAS$_{500}$ & $82.1\% \pm 0.8\%$ & $69.7\% \pm 1.4\%$ & $78.7\% \pm 0.6\%$ &  $\textbf{0.985} \pm 0.001$ \\
    \bottomrule
  \end{tabular*}
  \begin{flushleft}
    $^*$ We selected the best results reported by SGAT.
  \end{flushleft}
\end{table*}

\begin{table*}
  \caption{Link Prediction Results}
  \label{linkprediction}
  \begin{tabular*}{\linewidth}{l@{\extracolsep{\fill}}rrrrrr|rr}
    \toprule
    &
    \multicolumn{2}{c}{MVE\cite{qu2017mve}} &
    \multicolumn{2}{c}{MNE\cite{2018mne}} &
    \multicolumn{2}{c}{GATNE-T\cite{2019GATNE}} &
    \multicolumn{2}{c}{GATAS$_{100}$} \\
    \midrule
    Dataset & ROC-AUC & F1 & ROC-AUC & F1 & ROC-AUC & F1 & ROC-AUC & F1 \\
    \midrule
    Twitter$^*$ & 72.62 & 67.40 & 91.37 & 84.32 & 92.30 & 84.96 & \textbf{95.44} & \textbf{87.13} \\
    YouTube$^*$ & 70.39 & 65.10 & 82.30 & 75.03 & 84.61 & 76.83 & \textbf{96.63} & \textbf{83.59} \\
    \bottomrule
  \end{tabular*}
  \begin{flushleft}
    $^*$ Results reported for GATNE, MNE and MVE are from the original GATNE paper \cite{2019GATNE}.
  \end{flushleft}
\end{table*}

\subsection{Results}

Table \ref{results} summarizes our results on the node classification tasks. For the transductive tasks, we report the mean classification accuracy and standard deviation over 100 runs. For the inductive task, we report the mean micro-F1 score and standard deviation over 10 runs. We compare against the metrics already reported in \cite{velivckovic2017graph, kipf2016semi}, and use the same dataset splits provided. For GraphSAGE, we report the better results obtained in \cite{velivckovic2017graph}.

Using the settings described in the previous section, we provide variations of our method using different neighbourhood sample sizes: 10, 100, and 500. We notice that the model can achieve comparable performance, and that we have achieved a new state-of-the-art performance on the PPI dataset in an inductive setting, by a 1.2\% margin.

Performance increases with the neighbourhood sample size, as it expands the graph structure covered. However, we do not see substantial improvements for a sample size of 500. Given the average number of neighbours per node for each dataset, as shown in Table \ref{results}, we can see that an increased neighbourhood sample size of 500 might not add additional neighbours to the models in the transductive experiments. However, for the PPI dataset, 500 is still significantly below an estimated average neighbourhood size of $28.3^C$, where $C = 3$ is the number of steps considered.

We note that in the PPI dataset, a small neighbourhood sample size of 10 impacts performance considerably more than in the transductive setting. This could be because there is no support from the learnable node representations. As the amount of information provided by neighbours decreases, the model might become more dependent on these parameters. The proposed neighbourhood sampling technique trades in a small amount of accuracy to gain efficiency. As a result, the model can easily be used with large datasets and downstream tasks.

Table \ref{linkprediction} summarizes our results on the link prediction task. We report the macro area under the ROC curve and the macro F1 score for a single run. When comparing against GATNE, we use the transductive version of the model since we do not precompute raw features for the nodes and rely on the learned representations during test time. %Also, because the provided test split in \cite{2019GATNE} has low or no support for the majority of the edge types, we run experiments on an additional dataset using a node subset of the same size, generated by sampling nodes in the largest connected graph that have at least an edge of each type.

The results suggest the produced node representations are able to capture path attributes as part of the neighbourhood information. GATAS outperforms GATNE-T, with a lift of 3.14\% and 12.02\% in ROC-AUC, as well as 3.17\% and 6.76\% in the F1 score, on the Twitter and YouTube datasets respectively. The results achieved new state-of-the-art performances, to the best of our knowledge.

\subsection{Ablation Study}

The proposed architecture has three independent components that have not been considered in previous work: (1) the neighbour sampling technique using transition probabilities with learnable step coefficients that affect the attention weights; (2) the learnable node representations $\vec{\mathbf{l}}$ that augment node features with neighbourhood information; and (3) the attention network that allows neighbour representations to adapt to the target node given itself and the path information.

In this section we measure the impact of each component on the Cora and Pubmed datasets for the transductive setting and the PPI dataset for the inductive setting. We consider five model variations that test the importance of each component. All variations of the inductive models do not use learnable node representations because of the nature of its setting. We would like to test the importance of edge type information but the link prediction datasets do not provide node features that would allow us to run all variations. We define the following variations:

\begin{itemize}
  \item \textbf{\textit{Base}} only samples immediate neighbours with uniform probability and the transition probabilities are not part of the attention weights. The model does not adapt neighbour representations and nodes do not include learnable representations. Neighbour representations are reduced using the attention mechanism described in Section \ref{nodes}.
  \item \textbf{\textit{GATAS w/o trans}} is the proposed solution but all nodes within $C = 3$ steps can be sampled with uniform probability and the transition probabilities are not part of the attention weights.
  \item \textbf{\textit{GATAS w/o embed}} is the proposed solution but only node features are used such that $\vec{\mathbf{h}} = \vec{\mathbf{b}}$ and $\vec{\mathbf{l}}$ is not used. This variation is not available for the inductive task because of its nature.
  \item \textbf{\textit{GATAS w/o paths}} corresponds to the proposed solution but neighbour representations are not transformed according to the target node and path information.
  \item \textbf{\textit{GATAS}} is the proposed solution.
\end{itemize}

We ran experiments using the same settings described in Section \ref{experiment_setup}. In all the transductive variations, we use a sample size $S = 100$, since a larger value might attenuate the impact of using learnable representations, interfering with the results of the \textit{GATAS w/o embed} variation. For the variations in the inductive task we use a sample size of $S = 500$.

Table \ref{ablation} shows our results. The largest jump in performance corresponds to the use of the neighbourhood sampling technique and incorporation of the transition probabilities. The use of learnable embeddings as part of node representations does not seem to cause a big impact on performance but this could be a consequence of a large neighbourhood size $S$, which might reduce the need to utilize these parameters. Finally, the use of adaptable neighbour representations does not seem to affect performance for these tasks. We hypothesize that the nature of the tasks might not require such neighbour transformations but note that edge direction and different edge types are not present in these datasets.

\begin{table*}
	\caption{Ablation Study Results}
	\label{ablation}
	\begin{tabular*}{\linewidth}{l@{\extracolsep{\fill}}rrr}
		\toprule
		& \multicolumn{2}{c}{Transductive (Accuracy \%)} & \multicolumn{1}{c}{Inductive (Micro-F1)} \\
		\midrule
		Model & Cora & Pubmed & PPI \\
		\midrule
		Base & $79.3\% \pm 0.7\%$ & $76.2\% \pm 0.7\%$ & $0.974 \pm 0.001$ \\
    		GATAS w/o trans & $76.7\% \pm 0.8\%$ & $77.8\% \pm 0.5\%$ & $0.982 \pm 0.001$  \\
   		GATAS w/o embed & $82.2\% \pm 0.7\%$ & $78.3\% \pm 0.8\%$ & --- \\
		GATAS w/o paths & $82.3\% \pm 0.7\%$ & $78.4\% \pm 0.7\%$ & $0.983 \pm 0.001$  \\
		GATAS & $82.3\% \pm 0.9\%$ & $78.4\% \pm 0.6\%$ & $0.985 \pm 0.001$  \\
		\bottomrule
	\end{tabular*}
\end{table*}

\section{Conclusion}

In this paper, we proposed a new neural network architecture for graphs. The algorithm represents nodes by reducing their neighbour representations with attention. Multi-step neighbour representations incorporate different path properties. Neighbours are sampled using learnable depth coefficients.

Our model achieves comparable results across different tasks and various baselines, on the benchmark datasets: Cora, Citeseer, Pubmed, PPI, Twitter and YouTube. We successfully retained performance while increasing efficiency on large graphs, achieving state-of-the-art performance on multiple datasets from different tasks. We conducted an ablation study in transductive and inductive settings. The experiments show that sampling neighbourhoods according to weighted transition probabilities achieves the largest performance gain, especially in the inductive setting.

\bibliographystyle{plainnat}
\bibliography{neurips_2020}

\appendix
\newpage
\section{Supplementary Materials}

\subsection{Dataset Statistics}

\begin{table}[h]
	\caption{Dataset Statistics}
	\label{datastats}
  \begin{tabular*}{\linewidth}{l@{\extracolsep{\fill}}rrrrrr}
    \toprule
    &
    \textbf{Cora} &
    \textbf{Citeseer} &
    \textbf{PubMed} &
    \textbf{PPI} &
    \textbf{Twitter$^\dagger$} &
    \textbf{YouTube$^\dagger$} \\
    \midrule
    Node Classes & 7 & 6 & 3 & 121 & 1 & 1 \\
    Edge Types & 1 & 1 & 1 & 1 & 4 & 5 \\
    Node Features & 1,433 & 3,703 & 500 & 50 & 0 & 0 \\
    Nodes & 2,708 & 3,327 & 19,717 & 56,944 & 456,626 & 15,088 \\
    Edges & 5,429 & 4,732 & 44,338 & 818,716 & 15,367,315 & 13,628,895 \\
    Training Nodes & 140 & 120 & 60 & 44,906 & 9,990 & 2,000 \\
    Training Edges & ---$^*$ & ---$^*$ & ---$^*$ & 1,246,382 & 282,115 & 1,114,025 \\
    Validation Nodes & 500 & 500 & 500 & 6,514 & 9,891 & 2,000 \\
    Validation Edges & ---$^*$ & ---$^*$ & ---$^*$ & 201,647 & 16,463 & 65,512 \\
    Testing Nodes & 1,000 & 1,000 & 1,000 & 5,524 & 9,985 & 2,000 \\
    Testing Edges & ---$^*$ & ---$^*$ & ---$^*$ & 164,319 & 32,919 & 131,007 \\
    Neighbours per Node & 3.9 & 2.7 & 4.4 & 28.3 & 28.2 & 557.0 \\
    \bottomrule
  \end{tabular*}
  \begin{flushleft}
    $^\dagger$ Shared nodes across dataset types with access to training set edges only.
    \newline
    $^*$ Access to all edges.
  \end{flushleft}
\end{table}

\subsection{Experiment Settings}

\begin{table}[h]
	\caption{Experiment Settings}
	\label{parameters}
	\begin{tabular*}{\columnwidth}{l@{\extracolsep{\fill}}rrr}
    \toprule
    &
    \multicolumn{2}{c}{\textbf{Node Classification}} &
    \multicolumn{1}{c}{\textbf{Link Prediction}} \\
		\midrule
    \textbf{Parameter} &
    \textbf{Transductive} &
    \textbf{Inductive} &
    \textbf{Transductive} \\
		\midrule
    Maximum number of steps ($C$) & 3 & 3 & 2 \\
    Neighbourhood sample size ($S$) & 10/100/500 & 10/100/500 & 100 \\
    Layer size ($F$, $F'$, $F''$) & 50 &  50 & --- \\
    Node embedding size ($R$) & 10 & --- & 50 \\
    Edge type embedding size ($D$) & 10 & 5 & 50 \\
    Number of attention heads & 8 & 10 & 10 \\
    Input noise rate & 0.9 & 0 & 0 \\
    Dropout probability & 0.5 & 0 & 0 \\
    $L_2$ regularization coefficient ($\lambda$) & 0.05 & 0 & 0 \\
    Learning rate & 0.001 & 0.001 & 0.001 \\
    Maximum number of epochs & 1000 & 1000 & 1000 \\
    Early stopping patience & 100 & 10 & 5 \\
    Batch size ($B$) & 5000 & 100 & 200 \\
    \bottomrule
	\end{tabular*}
\end{table}

%References follow the acknowledgments. Use unnumbered first-level heading for
%the references. Any choice of citation style is acceptable as long as you are
%consistent. It is permissible to reduce the font size to \verb+small+ (9 point)
%when listing the references.
%{\bf Note that the Reference section does not count towards the eight pages of content that are allowed.}
%\medskip
%
%\small
%
%[1] Alexander, J.A.\ \& Mozer, M.C.\ (1995) Template-based algorithms for
%connectionist rule extraction. In G.\ Tesauro, D.S.\ Touretzky and T.K.\ Leen
%(eds.), {\it Advances in Neural Information Processing Systems 7},
%pp.\ 609--616. Cambridge, MA: MIT Press.
%
%[2] Bower, J.M.\ \& Beeman, D.\ (1995) {\it The Book of GENESIS: Exploring
%  Realistic Neural Models with the GEneral NEural SImulation System.}  New York:
%TELOS/Springer--Verlag.
%
%[3] Hasselmo, M.E., Schnell, E.\ \& Barkai, E.\ (1995) Dynamics of learning and
%recall at excitatory recurrent synapses and cholinergic modulation in rat
%hippocampal region CA3. {\it Journal of Neuroscience} {\bf 15}(7):5249-5262.
%
\end{document}